\documentclass[runningheads]{llncs}

\usepackage{eccv}

\usepackage{eccvabbrv}

\usepackage{graphicx}
\usepackage{multirow}
\usepackage{colortbl}
\usepackage{bm}
\usepackage{booktabs}
\usepackage{arydshln}  

\usepackage[accsupp]{axessibility}  

\usepackage{hyperref}

\usepackage{orcidlink}

\begin{document}

\title{Listen to Look into the Future: Audio-Visual Egocentric Gaze Anticipation} 

\titlerunning{Audio-Visual Egocentric Gaze Anticipation}

\author{Bolin Lai\inst{1} \and Fiona Ryan\inst{1} \and Wenqi Jia\inst{1} \and Miao Liu\inst{2,*} \and James M. Rehg\inst{3,*}}

\authorrunning{B. Lai et al.}

\institute{Georgia Institute of Technology \and
GenAI, Meta \and University of Illinois Urbana-Champaign \\
{\tt \{bolin.lai,fkryan,wenqi.jia\}@gatech.edu \quad miaoliu@meta.com \quad jrehg@illinois.edu}}

\maketitle

\let\thefootnote\relax\footnotetext{* Equal corresponding author.}

\vspace{-0.2cm}
\begin{abstract}
  Egocentric gaze anticipation serves as a key building block for the emerging capability of Augmented Reality. Notably, gaze behavior is driven by both visual cues and audio signals during daily activities. Motivated by this observation, we introduce the first model that leverages both the video and audio modalities for egocentric gaze anticipation. Specifically, we propose a Contrastive Spatial-Temporal Separable (CSTS) fusion approach that adopts two modules to separately capture audio-visual correlations in spatial and temporal dimensions, and applies a contrastive loss on the re-weighted audio-visual features from fusion modules for representation learning. We conduct extensive ablation studies and thorough analysis using two egocentric video datasets: Ego4D and Aria, to validate our model design. We demonstrate the audio improves the performance by +2.5\% and +2.4\% on the two datasets. Our model also outperforms the prior state-of-the-art methods by at least +1.9\% and +1.6\%. Moreover, we provide visualizations to show the gaze anticipation results and provide additional insights into audio-visual representation learning. The code and data split are available on our website (\url{https://bolinlai.github.io/CSTS-EgoGazeAnticipation/}).
  
\end{abstract}

\section{Introduction}
\label{sec:intro}

A person's eye movements during their daily activities are reflective of their intentions and goals (see~\cite{hayhoe2005eye} for a representative cognitive science study). The ability to predict the future gaze targets of the camera-wearer from egocentric videos, known as \emph{egocentric gaze anticipation}, is therefore a key step towards understanding and modeling cognitive processes and decision making. Furthermore, this capability could enable new applications in Augmented Reality and Wearable Computing, especially in social scenarios -- for example, providing memory aids for patients with cognitive impairments, or reducing the latency of content delivery in such AR systems. However, forecasting the gaze fixations of a camera-wearer using only the egocentric view (\ie, without eye tracking at testing time) is very challenging due to the complexity of egocentric scene content and the dynamic nature of gaze behaviors.

\begin{figure}[t]
\centering
\includegraphics[width=\linewidth]{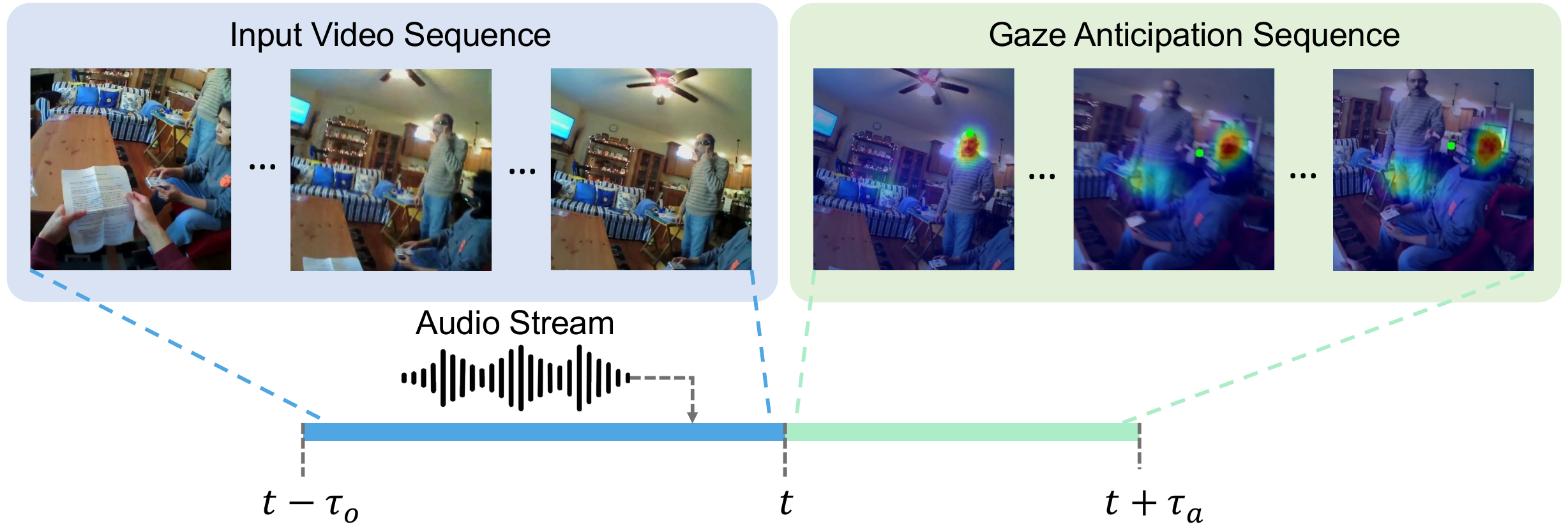}
\vspace{-0.4cm}
\caption{The problem setting of egocentric gaze anticipation. $\tau_o$ denotes the observation time, and $\tau_a$ denotes the anticipation time. Given the video frames and audio signals of the Input Video Sequence, the model seeks to predict the gaze fixation distribution for the time steps in the Gaze Anticipation Sequence. {\color{ForestGreen} Green dots} indicate the gaze targets in future frames and the heatmap shows the gaze anticipation result from our model. }
\vspace{-0.3cm}
\label{fig:teaser}
\end{figure}

We argue that audio signals can serve as an important auxiliary cue for egocentric gaze forecasting. Consider the example in \cref{fig:teaser}. In the input sequence, the camera view shifts from the paper held by the camera wearer to the standing speaker who asks a question. Then the sitting speaker on the far right answers the question, which is captured by the audio stream. In the anticipation sequence, the camera wearer's gaze shifts towards the sitting person's head after hearing her response. In this case, the audio stream (the sitting person's response) is an important stimulus that triggers this gaze movement. The influence of audio signals on eye movements is also evidenced by neuroscience research~\cite{schaefer1981acoustic}. Therefore, we address the problem of forecasting the gaze fixation of the camera-wearer in unseen future frames using a short egocentric video clip and corresponding audio. As shown in \cref{fig:teaser}, the model's ability to fuse the audio and visual cues enables it to correctly predict the future attention to the seated subject.

Though many works have addressed egocentric gaze estimation~\cite{huang2018predicting,huang2020ego,huang2020mutual,li2013learning,li2021eye,soo2015social,lai2022eye}, the egocentric gaze \textit{anticipation} task is largely understudied~\cite{zhang2017deep}. Moreover, how to leverage both the visual modality and the audio modality for egocentric gaze modeling has not been explored yet.Existing methods on audio-visual learning commonly~\cite{tsiami2020stavis,xiong2023casp,chang2021temporal,hu2020discriminative,qian2020multiple,arandjelovic2018objects,chen2021localizing} fuse visual and audio embeddings simultaneously in time and space. However, such a fusion mechanism is not ideal under the egocentric setting, where the camera wearer's reaction to the audio stimuli causes a drastic change of camera viewpoint. In \cref{fig:teaser}, as a reaction to the question and answer, the camera wearer shifts the attention from the paper to the standing person and then to the sitting person. The viewpoint and scene also have changed because of head movement (see the first and last frame). Moreover, due to the natural delay of reaction time, the audio stimulus and gaze reaction will not occur at the same time. Therefore, predicting the future gaze behavior demands a model can (1) learn possible viewpoint and scene change driven by the audio stream \emph{over time} and (2) locate the potential future gaze target \emph{in the visual space}. Fusing two modalities in time and space simultaneously may result in limited performance in the two targets because of spurious audio-visual correlations. Hence, a spatial-temporal separable fusion model is a better solution for egocentric gaze anticipation task.

To address the challenges in our task, we propose a novel \textbf{C}ontrastive \textbf{S}patial-\textbf{T}emporal \textbf{S}eparable (\textbf{CSTS}) audio-visual fusion method for egocentric gaze anticipation. Specifically, we input the egocentric video frames and the corresponding audio spectrograms into a video encoder and an audio encoder respectively. Then we develop a spatial fusion module and a temporal fusion module in parallel based on self-attention mechanism for modeling the spatial and temporal audio-visual correlation \emph{separately}, exactly addressing the aforementioned demands. The output representations from the two branches are merged by channel-wise reweighting and fed into a visual decoder to predict the future gaze target. We also propose a novel strategy that uses a multi-modal contrastive loss~\cite{akbari2021vatt} on the \emph{reweighted} representations (referred to as \emph{post-fusion} contrastive loss) from the fusion modules to facilitate audio-visual correspondence learning. We demonstrate the benefits of our approach on two egocentric video datasets that capture social scenarios and everyday activities: Ego4D~\cite{grauman2022ego4d} and Aria~\cite{aria_pilot_dataset}. The proposed model achieves state-of-the-art gaze anticipation performance on both datasets. Our contributions are summarized as follows:\vspace{-0.1em}
\vspace{-0.15cm}
\begin{itemize}
    \item[$\bullet$] We introduce the first approach that utilizes visual and audio signals for modeling egocentric gaze behaviors.
    \item[$\bullet$] We propose a novel CSTS model that leverages a spatio-temporal separable fusion module and a post-fusion contrastive learning scheme to facilitate audio-visual representation learning for egocentric gaze anticipation.
    \item[$\bullet$] We present comprehensive experiment results on the Ego4D~\cite{grauman2022ego4d} and Aria~\cite{aria_pilot_dataset} datasets. Our ablation studies show audio modality can improve the performance by +2.5\% and +2.4\% respectively in F1 score on Ego4D and Aria. The experiments also demonstrate our model outperforms prior state-of-the-art method by +1.9\% and +1.6\% in F1 score on the two datasets. 
\end{itemize}

\vspace{-0.3cm}
\section{Related Work}
\label{sec:related_work}
\vspace{-0.1cm}

\noindent\textbf{Egocentric Gaze Modeling}. Modeling human gaze behavior in egocentric videos is an important topic in egocentric vision. Most prior efforts target at egocentric gaze estimation \cite{li2021eye,lai2022eye,huang2018predicting,huang2020ego,li2013learning,huang2020mutual}. Huang \etal~\cite{huang2018predicting} propose learning temporal attention transitions from video features that reflect drastic gaze movements. Li \etal~\cite{li2021eye} and Huang \etal~\cite{huang2020mutual} utilize the correlation of gaze behaviors and actions, modeling them jointly with a convolutional network. Lai \etal~\cite{lai2022eye} encode global scene context into a single global token and explicitly model the global-local correlations in the visual embedding for gaze estimation. In contrast, egocentric gaze anticipation, which seeks to predict future gaze targets from past video frames, addresses an understudied dimension of modeling gaze. Zhang \etal \cite{zhang2017deep} introduce this task and utilize a convolutional network and a discriminator to generate future video frames, which are further used to anticipate future gaze targets. They enhance their model by adding an additional branch for gaze forecasting \cite{zhang2018anticipating}. All previous efforts on both egocentric gaze estimation and anticipation model gaze behavior from only the visual properties of the video stream, and do not consider the relationship between audio signals and gaze behavior. In this work, we introduce the first model that leverages both visual and audio signals for egocentric gaze anticipation task.

\noindent\textbf{Audio-Visual Saliency Prediction.} Audio-visual saliency prediction is a well-studied problem in computer vision~\cite{ruesch2008multimodal,schauerte2011multimodal,ratajczak2016fast,sidaty2017toward,coutrot2016multimodal,min2016fixation}. Another related research topic is sound source localization~\cite{arandjelovic2018objects,senocak2018learning,hu2019deep,hu2020discriminative,hu2022mix,huang2023egocentric} which localizes sound source in the image/video corresponding to a given audio stream. Here, we mainly discuss previous approaches for fusing audio and visual representations in saliency prediction problem. Early CNN-based approaches adopt a late-fusion strategy~\cite{tsiami2019behaviorally,tsiami2020stavis,min2020multimodal,wang2021semantic,wang2021weakly} for saliency prediction. Recently, new findings suggest audio-visual fusion at the intermediate features is a more effective way to leverage advantages of both modalities~\cite{agrawal2022does,yang2023svgc,cheng2021audio,tavakoli2019dave} for saliency prediction. Jain \etal~\cite{jain2021vinet} investigate two fusion methods at the middle level which achieve new state of the art on multiple datasets. Yao~\etal~\cite{yao2021deep} propose to incorporate the audio signal at multiple decoder layers by using an inner-product operation. Similarly, Chang~\etal~\cite{chang2021temporal} and Xiong~\etal~\cite{xiong2023casp} merge audio features into visual features at multiple levels of the visual encoder. Notably, our problem differs from the audio-visual saliency prediction in two aspects: First, the goal of our task is forecasting gaze behavior in the \emph{future}, while saliency prediction focuses more on studying human's attention mechanism in the \emph{current} video frame. Second, our problem focuses egocentric videos that capture the \emph{changing viewpoint} when people respond to audio and visual stimuli, while saliency prediction uses videos captured from a \emph{fixed} viewpoint, and fail to reflect gaze reaction to real-time events. Apart from the difference on problem settings, we also want to emphasize that the transformer-based fusion methods have not been applied in the audio-visual saliency prediction problem. Moreover, we propose a well-motivated spatio-temporal separable fusion module to address this challenging problem 

\noindent\textbf{Contrastive Audio-Visual Representation Learning}. Our work draws from a rich literature on leveraging contrastive learning to learn audiovisual feature representations \cite{arandjelovic2017look, korbar2018cooperative, morgado2021audio, morgado2020learning, morgado2021robust, patrick2020multi, ma2020active, alayrac2020self,akbari2021vatt,gong2022contrastive,gurram2022lava,ma2021contrastive}. These works learn correspondences between audio and visual signals in an self-supervised manner, constructing positive pairs from matching video frames and audio segments, and negative pairs from all other pairwise combinations. We employ a similar contrastive loss to learn correspondences between co-occurring audio and visual features. However, while prior methods calculate contrastive loss on the raw embedding from each modality, we propose to apply contrastive loss on re-weighted audio and visual representations from our proposed spatial and temporal fusion mechanism. 

\vspace{-0.1cm}
\section{Method}
\vspace{-0.1cm}
The egocentric gaze anticipation problem is illustrated in ~\cref{fig:teaser}. Given an egocentric video and audio from time $t-\tau_o$ to $t$, the goal is to predict the future gaze from $t$ to $t+\tau_a$ seconds. We denote the input video and audio as $x$ and $a$, respectively, and model the gaze fixation as a probabilistic distribution on a 2D image plane (following~\cite{li2021eye, lai2022eye}).

Notably, visual and audio signals have correlations in both spatial and temporal dimensions for gaze modeling. Spatially, the visual region (\eg, sound source) that has a stronger correlation with audio signals is more likely to be the potential future gaze target. Temporally, events in the audio signal may drive both egocentric viewpoint change (via head movement) and gaze movements as the camera wearer responds to new sounds. Our key insight is thus separating the connection of audio and visual signals into spatial and temporal correlations.

\begin{figure}[t]
\centering
\includegraphics[width=\linewidth]{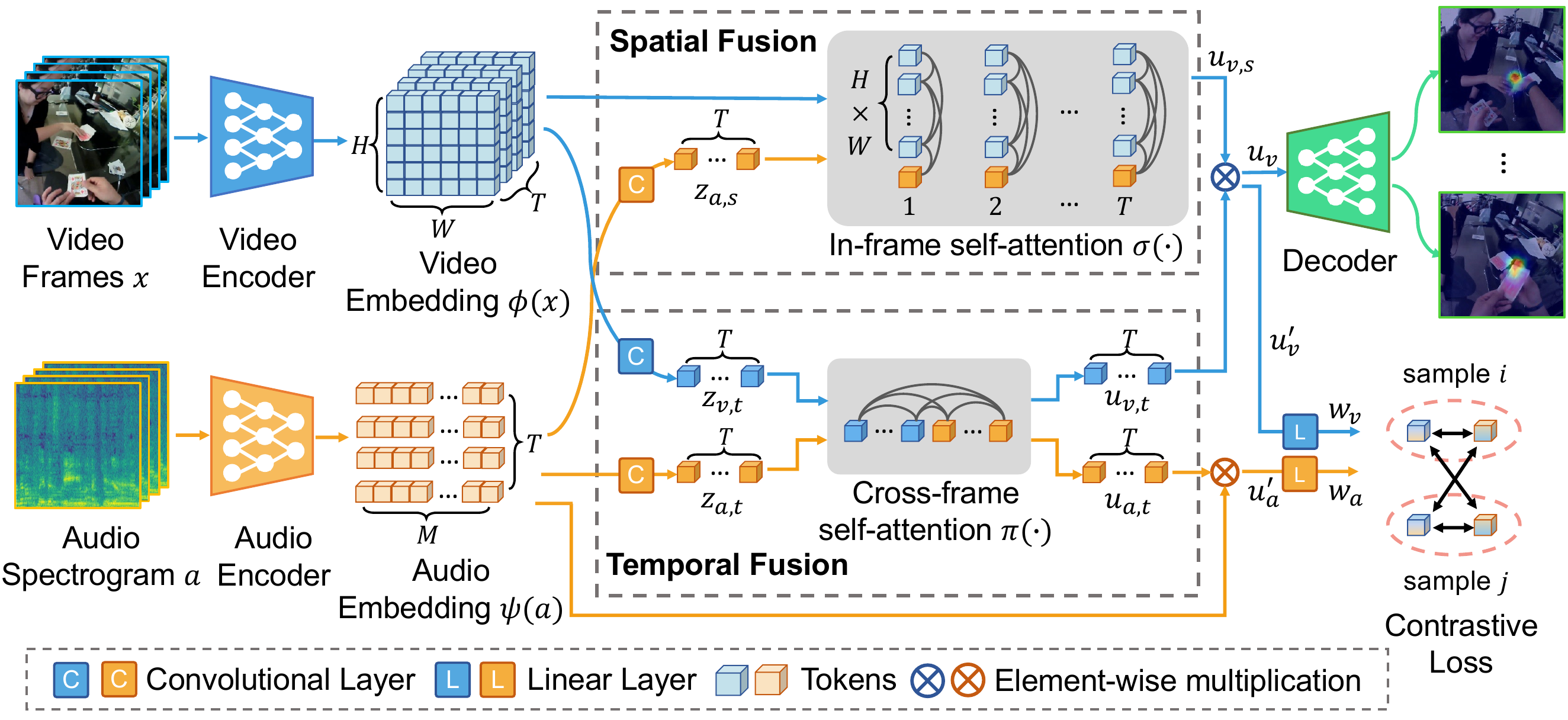}
\vspace{-0.1cm}
\caption{Overview of the proposed model. The video embeddings $\phi(x)$ and audio embeddings $\psi(a)$ are obtained by two transformer-based encoders. We then model the correlations of visual and audio embeddings using two separate branches -- (1) spatial fusion, which learns the spatial co-occurence of audio signals and visual objects in each frame, and (2) temporal fusion, which captures the temporal correlations and possible gaze movement. A contrastive loss is adopted to facilitate audio-visual representation learning. We input fused embeddings into a decoder for final gaze anticipation results.}
\vspace{-0.2cm}
\label{fig:overview}
\end{figure}

\cref{fig:overview} demonstrates the overview of our model. We exploit the transformer-based encoders $\phi(x)$ and $\psi(a)$ to extract the representations of the video frames $x$ and audio signals $a$. We then employ a \textbf{C}ontrastive \textbf{S}patial-\textbf{T}emporal \textbf{S}eparable  (CSTS) audio-visual fusion approach. Specifically, a spatial fusion module captures the correlation between audio embeddings and spatial appearance-based features; a temporal fusion module captures the temporal correlation between the visual and audio embeddings; and a contrastive loss is applied on fused audio-visual embeddings to facilitate the representation learning. Finally, spatially and temporally fused audio-visual features are merged and fed into a decoder for future gaze anticipation.

\vspace{-0.2cm}
\subsection{Audio and Visual Feature Embedding}

\noindent\textbf{Visual Feature Embedding}. We adopt the multi-scale vision transformer (MViT) architecture~\cite{fan2021multiscale} as the video encoder $\phi(x)$. $\phi(x)$ splits the 3D video tensor input into multiple non-overlapping patches, and thereby extracts $T\times H\times W$ visual tokens with feature dimension $D$ from $x$.

\noindent\textbf{Audio Feature Embedding}. We follow~\cite{kazakos2019epic} to adopt a sliding window approach for audio signal preprocessing. Specifically, for a video frame at time step $t_i$, the corresponding audio segment has a range of $[t_i-\frac{1}{2}\Delta t_w, t_i+\frac{1}{2}\Delta t_w]$. We then use STFT to convert all audio segments into log-spectrograms and feed the processed audio segments into a transformer-based audio encoder $\psi(a)$. Since the audio stream has more sparse information than video stream, we adopt a light-weighted transformer architecture (inspired by~\cite{gao2020listen,dosovitskiy2020image}) for the audio encoder $\psi(a)$. In this way, $\psi(a)$ extracts $T\times M$ tokens with feature dimension $D$ from the audio inputs $a$.

\subsection{Spatial-Temporal Separable Fusion}
\noindent\textbf{Spatial Audio-Visual Fusion}.\ The spatial fusion branch identifies correlations between the audio signal corresponding to a video frame and its visual content in space. We first use convolutional operations to generate the audio representation $z_{a,s}$ for spatial fusion with dimensions $T\times 1\times D$ from the audio embedding $\psi(a)$. This allows the model to extract a holistic audio embedding within each sliding window. We then input the visual embedding $\phi(x)$ and pooled audio embedding $z_{a,s}$ into an in-frame self-attention layer $\sigma$. In this layer, we masked out all cross-frame connections and only calculate the correlations among visual tokens within each frame and the corresponding single audio token. Therefore, the input to the spatial fusion consists of $T$ groups of visual tokens, and $T$ single audio embeddings. Formally, we have:
\begin{equation}\label{eq:notion-spatial-v}
\phi(x)=\left[\phi(x)^{(1)}, ..., \phi(x)^{(T)}\right],
\end{equation}
\begin{equation}\label{eq:notion-spatial-a}
\vspace{-0.1cm}
z_{a,s}=\left[z_{a,s}^{(1)},...,z_{a,s}^{(T)}\right],
\end{equation}

\noindent where $\phi(x)^{(i)} \in\mathbb{R}^{1\times N\times D}$, $z_{a,s}^{(i)}\in\mathbb{R}^{1\times1\times D}$ with $i\in\{1,...,T\}$, and $N = H\times W$. Hence, the input from each time step is denoted as:
\begin{equation}\label{eq:notion-spatial-aa}
    z_s^{(i)}=\left[\phi^{(i)}(x),z_{a,s}^{(i)}\right]\in\mathbb{R}^{1\times(N+1)\times D}
\end{equation}
The in-frame self-attention operation for time step $i$ can be written as:
\begin{equation}\label{eq:in-frame}
    \sigma(z_s^{(i)}) = Softmax\left(\bm{Q}_s^{(i)}{\bm{K}_s^{(i)}}^T/\sqrt{D}\right)\bm{V}_s^{(i)}\in\mathbb{R}^{1\times(N+1)\times D},
\end{equation}
where $\bm{Q_s^{(i)}}, \bm{K}_s^{(i)}, \bm{V}_s^{(i)}$ refer to query, key, and value of the spatial self-attention at time step $i$, respectively. We apply \cref{eq:in-frame} independently for each time step $i$ and have the following overall in-frame self-attention:  
\begin{equation}
\sigma(z_s)=\left[\sigma(z_s^{(i)}),...,\sigma(z_s^{(T)})\right]\in\mathbb{R}^{T\times(N+1)\times D}.
\end{equation}
In practice, we input all tokens to the in-frame self-attention layer simultaneously, mask out cross-frame correlations and calculate \cref{eq:in-frame} in one shot to speed up training. We further add two linear layers after the self-attention outputs $\sigma(z_s)$, following the standard self-attention layer design. The output of the spatial module is finally denoted as $u_s\in\mathbb{R}^{T\times(N+1)\times D}$.

\noindent\textbf{Temporal Audio-Visual Fusion}.\ The temporal fusion 
 branch models relationships between audio and visual content across time. We apply two convolutional layers to integrate the embedding from each modality at each time step into a single token. The resulting visual and audio tokens are denoted as $z_{v,t}\in\mathbb{R}^{T\times1\times D}$ and $z_{a,t}\in\mathbb{R}^{T\times1\times D}$, respectively. Then we feed $z_t=\left[z_{v,t},z_{a,t}\right]\in\mathbb{R}^{2T\times1\times D}$ into a cross-frame self-attention layer $\pi$ that can be formulated as:
\begin{equation}
    \pi(z_t) = Softmax\left(\bm{Q}_t\bm{K}_t^T/\sqrt{D}\right)\bm{V}_t\in\mathbb{R}^{2T\times1\times D},
\end{equation}
where $\bm{Q}_t,\bm{K}_t,\bm{V}_t$ are query, key and value matrices with dimension $2T\times1\times D$. Similar to the spatial fusion, two additional linear layers are added after $\pi(z_t)$ and result in the final temporal fusion output $u_t\in\mathbb{R}^{2T\times1\times D}$.

\noindent\textbf{Merging of Two Fusion Modules}.\ After obtaining audio-visual representations from the two fusion modules, we merge the two branches by reweighting the output from spatial fusion with the temporal weights from temporal fusion in each channel, which produces a new representation for each modality that has been refined by multimodal spatial and temporal correlation. Specifically, we break down the output from spatial fusion $u_s\in\mathbb{R}^{T\times(N+1)\times D}$ into $u_{v,s}\in\mathbb{R}^{T\times N\times D}$ and $u_{a,s}\in\mathbb{R}^{T\times1\times D}$, and the output from temporal fusion $u_t\in\mathbb{R}^{2T\times1\times D}$ into $u_{v,t}\in\mathbb{R}^{T\times1\times D}$ and $u_{a,t}\in\mathbb{R}^{T\times1\times D}$. The reweighted visual representation is formulated as
\begin{equation}\label{eq:reweight}
\vspace{-0.04cm}
    u_v = u_{v,s} \otimes u_{v,t} \in\mathbb{R}^{T\times N\times D},
\vspace{-0.04cm}
\end{equation}
where $\otimes$ denotes element-wise multiplication with broadcast mechanism. $u_v$ is then fed into a decoder to generate final prediction for future gaze target. We follow~\cite{lai2022eye} to add skip connections from the video encoder to the decoder and optimize the network with a KL-Divergence loss $\mathcal{L}_{kld}$.

\vspace{-0.1cm}
\subsection{Contrastive Learning for Audio-Visual Fusion}

In addition to using KL-Divergence loss to supervise gaze anticipation, we propose to leverage the intrinsic alignment of visual and audio modalities to learn a more robust audio-visual representation by using a contrastive learning scheme. Multi-modal contrastive loss has been proved to be effective in self-supervised learning~\cite{alayrac2020self,akbari2021vatt}. Rather than calculating the contrastive loss directly on the raw embedded features, we innovatively propose to use the reweighted video and audio representations from the spatial and temporal fusion modules, which has not been studied in prior works. In our experiments, we show this is a more effective representation learning method for egocentric gaze anticipation.

To this end, we reweight the raw audio embedding $\psi(a)\in\mathbb{R}^{T\times M\times D}$ from the audio encoder by temporal weights $u_{a,t}$ from the temporal fusion module in a similar way to \cref{eq:reweight}. We then get the reweighted audio feature as

\begin{equation}
    u_a = \psi(a)\otimes u_{a,t} \in\mathbb{R}^{T\times M\times D}
\end{equation}
We don't use an additional learnable token to aggregate information from other tokens as prior works did~\cite{alayrac2020self,akbari2021vatt,lin2022egocentric}. We instead average all tokens of ${u_v}$ and ${u_a}$ respectively to obtain the single-vector representations $u'_v, u'_a\in\mathbb{R}^{1\times D}$ and then map them to a low-dimensional common space using linear layers followed by L2 normalization. It can be formulated as $w_v = Norm\left(f_1(u'_v)\right)$ and $w_a = Norm\left(f_2(u'_a)\right)$, where $f_1(\cdot),f_2(\cdot)$ are linear layers. The resulting visual vector and audio vector are denoted as $w_v,w_a\in\mathbb{R}^{1\times D'}$, where $D'$ is the new dimension of the common space. Within each mini-batch, corresponding audio and visual embeddings are considered as positive pairs, and all other pairwise combinations are considered as negative. Following~\cite{lin2022egocentric}, we calculate video-to-audio loss and audio-to-video loss separately. The video-to-audio contrastive loss is defined as
\begin{equation}
    \mathcal{L}^{v2a}_{cntr} = -\frac{1}{|\mathcal{B}|}\sum_{i=1}^{|\mathcal{B}|}\log\frac{\exp({w_v^{(i)}}^T w_a^{(i)}/\mathcal{T})}{\sum_{j\in\mathcal{B}}\exp({w_v^{(i)}}^T w_a^{(j)}/\mathcal{T})},
\end{equation}
where $\mathcal{B}$ is the training batch $\mathcal{B}=\{1,2,\dots,n\}$ and $\mathcal{T}$ is the temperature factor. Superscripts $(i)$ and $(j)$ denote the $i$-th and $j$-th samples in the batch. The audio-to-video loss is defined in a symmetric way. Finally, the contrastive loss is defined as $\mathcal{L}_{cntr}=\mathcal{L}_{cntr}^{v2a}+ \mathcal{L}_{cntr}^{a2v}$. $\mathcal{L}_{kld}$ and $\mathcal{L}_{cntr}$ are linearly combined with a parameter $\alpha$ for the final training loss, \ie, $\mathcal{L} = \mathcal{L}_{kld} + \alpha\mathcal{L}_{cntr}$.

\subsection{Implementation Details}
In our experiments, we set observation time $\tau_o$ as $3$ seconds and anticipation time $\tau_a$ as 
$2$ seconds. For video input, we sample 8 frames from the observable segment and resize to a spatial size of 256$\times$256. For audio input, following~\cite{kazakos2019epic}, we first resample the audio signal to 24kHz and set a time window with $\Delta t_w =1.28s$ to crop the audio segment corresponding to each video frame. We then convert it to a log-spectrogram using a STFT with window size 10ms and hop length 5ms. The number of frequency bands is set as 256 resulting in a spectrogram matrix of size 256$\times$256. The output of the decoder is the gaze distribution on 8 frames uniformly sampled from the 2-second anticipation time. More details about model architecture and training hyper-parameters can be found in supplementary.

\section{Experiments}

We first introduce the datasets and evaluation metrics used in our experiments. We then present detailed ablation studies to validate the contribution of each component in our method, and demonstrate the performance improvement over prior state-of-the-art methods for gaze anticipation as well as gaze estimation models applied to the gaze anticipation task. Finally, we visualize the predictions and weights of our model to provide qualitative insight into our method.

\subsection{Experiment Setup}
\noindent\textbf{Datasets}.\ We conduct experiments on two egocentric datasets that contain aligned video and audio streams and gaze tracking data -- Ego4D~\cite{grauman2022ego4d} and Aria~\cite{aria_pilot_dataset}. Note that another widely used gaze estimation benchmark EGTEA Gaze+~\cite{li2021eye} does not release audio data and thus is not feasible for our study. Other popular egocentric video datasets, such as Epic-Kitchens~\cite{damen2018scaling} and Charades-Ego~\cite{sigurdsson2018charades}, are also not applicable to our task because they don't have eye-tracking labels. Ego4D and Aria are the two largest public datasets that provide all necessary data and labels (\ie, egocentric videos, aligned audio streams and eye-tracking data) for egocentric audio-visual gaze anticipation.

The Ego4D eye-tracking subset is collected in social settings (\ie, social interaction benchmark) and totals 31 hours of egocentric videos from 80 participants. All videos have a fixed 30 fps frame rate and spatial resolution of 1088$\times$1080, and audio streams are recorded with a sampling rate of 44.1kHz. We use the train/test split released in~\cite{lai2022eye} in our experiments, \ie, 15310 video segments for training and the other 5202 video segments for testing.

The Aria dataset contains 143 egocentric videos (totaling 7.3 hours) collected with Project Aria glasses. It covers a variety of indoor everyday activities including cooking, exercising and spending time with friends. All videos have a fixed 20 fps frame rate and spatial resolution of 1408$\times$1408. A sliding window is used to trim long videos into 5-second video segment with a stride of 2 seconds. We use 107 videos (10456 segments) for training and 36 videos (2901 segments) for testing. We will release our split to facilitate future studies in this direction.

\noindent \textbf{Evaluation Metrics.} As suggested in recent work on egocentric gaze estimation~\cite{lai2022eye}, AUC score can easily get saturated due to the long-tailed distribution of gaze on 2D video frames. Therefore, we follow~\cite{lai2022eye,li2021eye} to adopt F1 score (\emph{primary}), recall and precision as our evaluation metrics.

\subsection{Ablation Study}
\label{sec:ablation}

We first quantify the performance contribution of each key module from our proposed method. Specifically, we denote the model only using our proposed spatial fusion module as \emph{S-fusion}, the model only using our proposed temporal fusion module as \emph{T-fusion}, the model using both modules and our spatial-temporal separable fusion strategy \emph{without} the contrastive learning schema as \emph{STS}. We finally present the performance of our full CSTS model (\ie, \emph{STS} + contrastive learning). As demonstrated in \cref{tab:ablation}, compared with models trained solely on RGB frames (Vision only), S-fusion and T-fusion boost the F1 score by +1.4\% and +1.5\% on Ego4D, and +1.1\% and +1.1\% on Aria. Moreover, the STS model further achieves a F1 score of 39.2\% on Ego4D and 59.3\% on Aria. These results suggest that both the spatial and and the temporal correlation between video and audio signal play a vital role for egocentric gaze anticipation. Contrastive loss further improves F1 score by +0.5\% and +0.6\% suggesting its contributions to audio-visual representative learning. We also observe that the full model doesn't achieve the best in recall. This is because some incomplete baselines don't leverage audio modality as effectively as the full model and thus produce more uncertainty in output, resulting in higher recall and lower precision. Therefore, we consider F1 as the \emph{primary} metric. Similar phenomenon is also observed in the following experiments.

\begin{table}[t]
\caption{Ablations on each key component of our proposed model. \emph{CSTS} (highlighted in {\color{ForestGreen} green}) refers to the complete model of our approach. The best results are highlighted with \textbf{boldface}. Please refer to \cref{sec:ablation} for more discussions.}
\centering
\begin{tabular}{lcccccc}
\toprule
\makebox[0.15\textwidth][l]{\multirow{2}{*}{Methods}}  & \multicolumn{3}{c}{Ego4D} & \multicolumn{3}{c}{Aria} \\
\cmidrule(lr){2-4} \cmidrule(lr){5-7}
& \makebox[0.11\textwidth][c]{F1 Score} & \makebox[0.11\textwidth][c]{Recall} & \makebox[0.11\textwidth][c]{Precision} & \makebox[0.11\textwidth][c]{F1 Score} & \makebox[0.11\textwidth][c]{Recall} & \makebox[0.11\textwidth][c]{Precision} \\
\midrule
Vision only  & 37.2 & 54.1 & 28.3 & 57.5 & 62.4 & 53.3 \\
S-fusion     & 38.6 & \textbf{54.1} & 30.1 & 58.6 & \textbf{67.1} & 52.0 \\
T-fusion     & 38.7 & 53.8 & 30.1 & 58.6 & 65.9 & 52.8 \\
STS          & 39.2 & 53.7 & 30.8 & 59.3 & 66.8 & 53.3 \\
\rowcolor[HTML]{cef8d1} CSTS  & \textbf{39.7} & 53.3 & \textbf{31.6} & \textbf{59.9} & 66.8 & \textbf{54.3} \\
\bottomrule
\end{tabular}
\label{tab:ablation}
\vspace{-0.1cm}
\end{table}

\begin{table}[t]
\caption{Analysis on proposed fusion strategies. The best results are highlighted with \textbf{boldface}. \emph{STS} (highlighted in {\color{ForestGreen} green}) refers to the proposed spatial-temporal separable fusion method (without contrastive learning). More discussions are in \cref{sec:analysis}.}
\centering
\begin{tabular}{lcccccc}
\toprule
\makebox[0.15\textwidth][l]{\multirow{2}{*}{Methods}}  & \multicolumn{3}{c}{Ego4D} & \multicolumn{3}{c}{Aria} \\
\cmidrule(lr){2-4} \cmidrule(lr){5-7}
& \makebox[0.11\textwidth][c]{F1 Score} & \makebox[0.11\textwidth][c]{Recall} & \makebox[0.11\textwidth][c]{Precision} & \makebox[0.11\textwidth][c]{F1 Score} & \makebox[0.11\textwidth][c]{Recall} & \makebox[0.11\textwidth][c]{Precision} \\
\midrule
Vision only     & 37.2 & \textbf{54.1} & 28.3 & 57.5 & 62.4 & 53.3 \\
Linear          & 38.2 & 53.0 & 29.9 & 58.1 & 65.9 & 51.9 \\
Bilinear        & 37.6 & 52.8 & 29.2 & 57.7 & 66.8 & 50.8 \\
Concat.         & 38.1 & 53.6 & 29.5 & 58.0 & 66.8 & 51.2 \\
Vanilla SA      & 38.5 & 53.3 & 30.1 & 58.0 & 67.2 & 51.1 \\
\rowcolor[HTML]{cef8d1} STS  & \textbf{39.2} & 53.7 & \textbf{30.8} & \textbf{59.3} & \textbf{66.8} & \textbf{53.3} \\
\bottomrule
\end{tabular}
\label{tab:other_fusion}
\vspace{-0.1cm}
\end{table}

\subsection{Analysis on Fusion and Contrastive Learning Strategies}
\label{sec:analysis}

Directly feeding all visual and audio tokens into a fusion layer (\ie, joint fusion) is a widely used approach for audio-visual saliency prediction~\cite{tsiami2020stavis,xiong2023casp,chang2021temporal} and action recognition~\cite{kazakos2019epic,gao2020listen,wang2020makes}. To show the superiority of the proposed spatial-temporal separable (\emph{STS}) fusion approach in handling the unique challenges of our task, we provide additional comparison with four joint fusion strategies that are widely used in audio-visual saliency prediction and audio-visual action recognition. Specifically, the four strategies are (1) fusing two modalities with a few linear layers~\cite{gao2020listen} (denoted as \emph{Linear}); (2) feeding video and audio embeddings to a single bilinear layer~\cite{jain2021vinet,yao2021deep} (denoted as \emph{Bilinear}); (3) concatenating audio and visual embeddings along channel dimension (denoted as \emph{Concat.}) as in~\cite{kazakos2019epic,jain2021vinet}; (4) feeding all embedded video and audio tokens together into a standard self-attention layer (denoted as \emph{Vanilla SA}), inspired by~\cite{lin2023vision,xiong2023casp}. We replace our fusion modules with the four strategies in our framework for a fair comparison. We elaborate the implementation details of each baseline in supplementary.

As shown in \cref{tab:other_fusion}, Linear, Bilinear, Concat. and Vanilla SA methods have limited improvement over the vision-only baseline, suggesting that previous fusion strategies for audio-visual saliency prediction and general action recognition are sub-optimal for our problem setting. In contrast, our proposed fusion strategy (STS) yields larger performance boost (+2.0\% on Ego4D and +1.8\% on Aria) even without using the contrastive loss, which shows the benefits of spatial-temporal separable fusion mechanism. The possible reason is that prior joint fusion methods are designed for third-person videos without a drastic viewpoint change. However, forecasting gaze in egocentric view has the unique challenges caused by camera movement and the latency of gaze response to audio stimuli. Our approach fuses two modalities in space and time separately and hence avoids spurious correlations that may happen in joint fusion baselines.

\begin{table}[t]
\caption{Analysis on the proposed contrastive learning schema. \emph{Post Contr} denotes our post-fusion contrastive learning. \emph{STS + Post Contr} refers to the complete CSTS model. The best results are highlighted with \textbf{boldface}. More discussions are in \cref{sec:analysis}.}
\centering
\begin{tabular}{lcccccc}
\toprule
\makebox[0.2\textwidth][l]{\multirow{2}{*}{Methods}} & \multicolumn{3}{c}{Ego4D} & \multicolumn{3}{c}{Aria} \\
\cmidrule(lr){2-4} \cmidrule(lr){5-7}
& \makebox[0.1\textwidth][c]{F1 Score} & \makebox[0.1\textwidth][c]{Recall} & \makebox[0.1\textwidth][c]{Precision} & \makebox[0.1\textwidth][c]{F1 Score} & \makebox[0.1\textwidth][c]{Recall} & \makebox[0.1\textwidth][c]{Precision} \\
\midrule
Vanilla SA            & 38.5 & 53.3 & 30.1 & 58.0 & \textbf{67.2} & 51.1 \\
SA + Vanilla Contr    & 38.5 & 52.4 & 30.5 & 58.4 & 67.0 & 51.8 \\
SA + Post Contr       & 38.9 & \textbf{54.4} & 30.3 & 58.8 & 66.4 & 52.8 \\
\hdashline
STS                   & 39.2 & 53.7 & 30.8 & 59.3 & 66.8 & 53.3 \\
STS + Vanilla Contr   & 39.0 & 53.7 & 30.6 & 59.1 & 66.5 & 53.1 \\
\rowcolor[HTML]{cef8d1} STS + Post Contr  & \textbf{39.7} & 53.3 & \textbf{31.6} & \textbf{59.9} & 66.8 & \textbf{54.3} \\
\bottomrule
\end{tabular}
\label{tab:contr}
\vspace{-0.2cm}
\end{table}

We also evaluate the benefits of our proposed post-fusion contrastive learning scheme in \cref{tab:contr}. Here, we consider another baseline (denoted as \emph{Vanilla Contr}) that calculates the contrastive loss using raw video and audio embeddings (\ie, $\phi(x)$ and $\psi(a)$ in \cref{fig:overview}), as is typical in prior work~\cite{Tian_2018_ECCV,ma2020active,gong2022contrastive,gurram2022lava}. Our novel strategy of adding contrastive loss on fused features is denoted as \emph{Post Contr}. Vanilla Contr makes only minor differences on Vanilla SA model and even slightly reduces performance when accompanied by our proposed STS mechanism. In contrast, our proposed Post Contr scheme improves the performance of Vanilla SA by +0.4\% and 0.8\% and improves STS by +0.5\% and +0.6\% on the two datasets. These results further suggest that post-fusion contrastive learning is more robust for audio-visual learning in our task. More experiments of different contrastive learning strategies are provided in supplementary.

\vspace{-0.1cm}
\subsection{Comparison with State-of-the-art Methods}
\label{sec:sota}

\begin{table}[t]
\caption{Comparison with state-of-the-art models on egocentric gaze anticipation. We also adapt previous egocentric gaze estimation approaches to the anticipation setting for a more thorough comparison. The best results are highlighted with \textbf{boldface}. The \textcolor{ForestGreen}{green} row shows our model performance. Please refer to \cref{sec:sota} for more discussions.}
\vspace{-0.1cm}
\centering
\begin{tabular}{lcccccc}
\toprule
\makebox[0.17\textwidth][l]{\multirow{2}{*}{Methods}} & \multicolumn{3}{c}{Ego4D} & \multicolumn{3}{c}{Aria} \\
\cmidrule(lr){2-4} \cmidrule(lr){5-7}
& \makebox[0.11\textwidth][c]{F1 Score} & \makebox[0.11\textwidth][c]{Recall} & \makebox[0.11\textwidth][c]{Precision} & \makebox[0.11\textwidth][c]{F1 Score} & \makebox[0.11\textwidth][c]{Recall} & \makebox[0.11\textwidth][c]{Precision} \\
\midrule
Center Prior & 13.6 & 9.4 & 24.2 & 24.9 & 17.3 & 44.4 \\
GazeMLE~\cite{li2021eye} & 36.3 & 52.5 & 27.8 & 56.8 & 64.1 & 51.0 \\
AttnTrans~\cite{huang2018predicting} & 37.0 & \textbf{55.0} & 27.9 & 57.4 & 65.5 & 51.0 \\
I3D-R50~\cite{feichtenhofer2019slowfast} & 36.9 & 52.1 & 28.6 & 57.4 & 63.6 & 52.2 \\
MViT~\cite{fan2021multiscale} & 37.2 & 54.1 & 28.3 & 57.5 & 62.4 & 53.3 \\
GLC~\cite{lai2022eye} & 37.8 & 52.9 & 29.4 & 58.3 & 65.4 & 52.6 \\\hdashline
DFG~\cite{zhang2017deep}  & 37.2 & 53.2 & 28.6 & 57.4 & 63.6 & 52.3 \\
DFG+~\cite{zhang2018anticipating}  & 37.3 & 52.3 & 29.0 & 57.6 & 65.5 & 51.3 \\
\rowcolor[HTML]{cef8d1} CSTS & \textbf{39.7} & 53.3 & \textbf{31.6} & \textbf{59.9} & \textbf{66.8} & \textbf{54.3} \\
\bottomrule
\end{tabular}
\label{tab:sota_forecast}
\vspace{-0.3cm}
\end{table}

\begin{figure}[t]
\centering
\includegraphics[width=\linewidth]{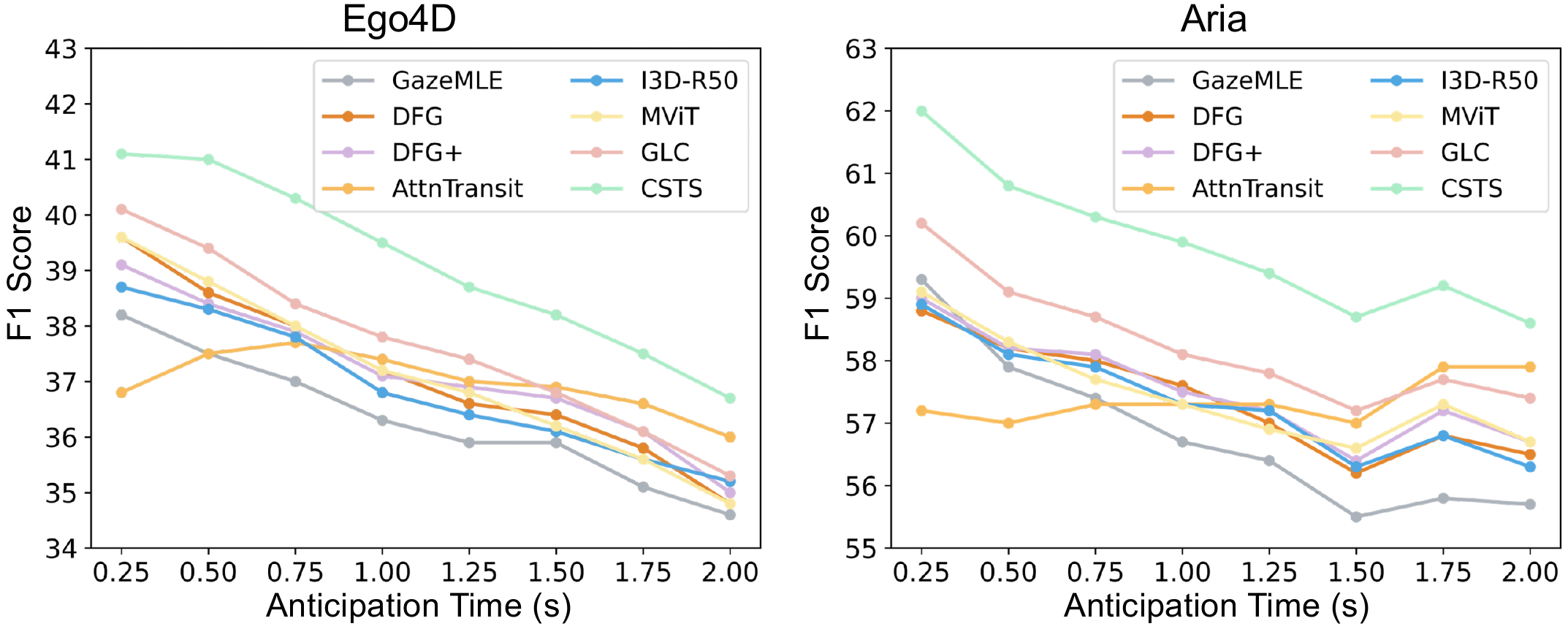}
\caption{The performance of gaze anticipation in each frame. Our model (CSTS) consistently outperforms all prior methods by a notable margin.}
\label{fig:perframe}
\vspace{-0.2cm}
\end{figure}

Most existing works on egocentric gaze modeling target at egocentric gaze estimation rather than anticipation. In order to provide a thorough comparison, in addition to comparing against SOTA egocentric gaze anticipation models (DFG\cite{zhang2017deep}, DFG+\cite{zhang2018anticipating}), we also adapt the recent SOTA egocentric gaze estimation model GLC\cite{lai2022eye} and all baselines from \cite{lai2022eye} (I3D-Res50 \cite{wang2018non}, MViT\cite{fan2021multiscale}, GazeMLE\cite{li2021eye} and AttnTrans\cite{huang2018predicting}) to the anticipation task.

As presented in \cref{tab:sota_forecast}, our method outperforms its direct competitor DFG+, which is the previous SOTA model for egocentric gaze anticipation, by +2.4\% F1 on Ego4D and +2.3\% F1 on Aria. Note that the original DFG and DFG+ used a less powerful backbone encoder, so for fair comparison, we reimplement their method using the same MViT backbone as our method. We also observe that methods originally designed for egocentric gaze estimation still work as strong baselines for the egocentric gaze anticipation task. Our proposed CSTS model also outperforms these methods, surpassing the recent SOTA for egocentric gaze estimation -- GLC by +1.9\% F1 on Ego4D and +1.6\% F1 on Aria. In addition, We also incorporate audio stream into the strongest baseline (GLC) by a straightforward concatenation whose F1 score is 38.1\% on Eg4D  and 58.5\% on Aria. The marginal gain over GLC (+0.3\%/+0.2\%) suggests that simply using audio stream in a strong baseline without specific design leads to sub-optimal solution in egocentric gaze anticipation problem, which in turn validates the effectiveness and necessity of our approach.

\begin{figure}[t]
\begin{center}
\includegraphics[width=\linewidth]{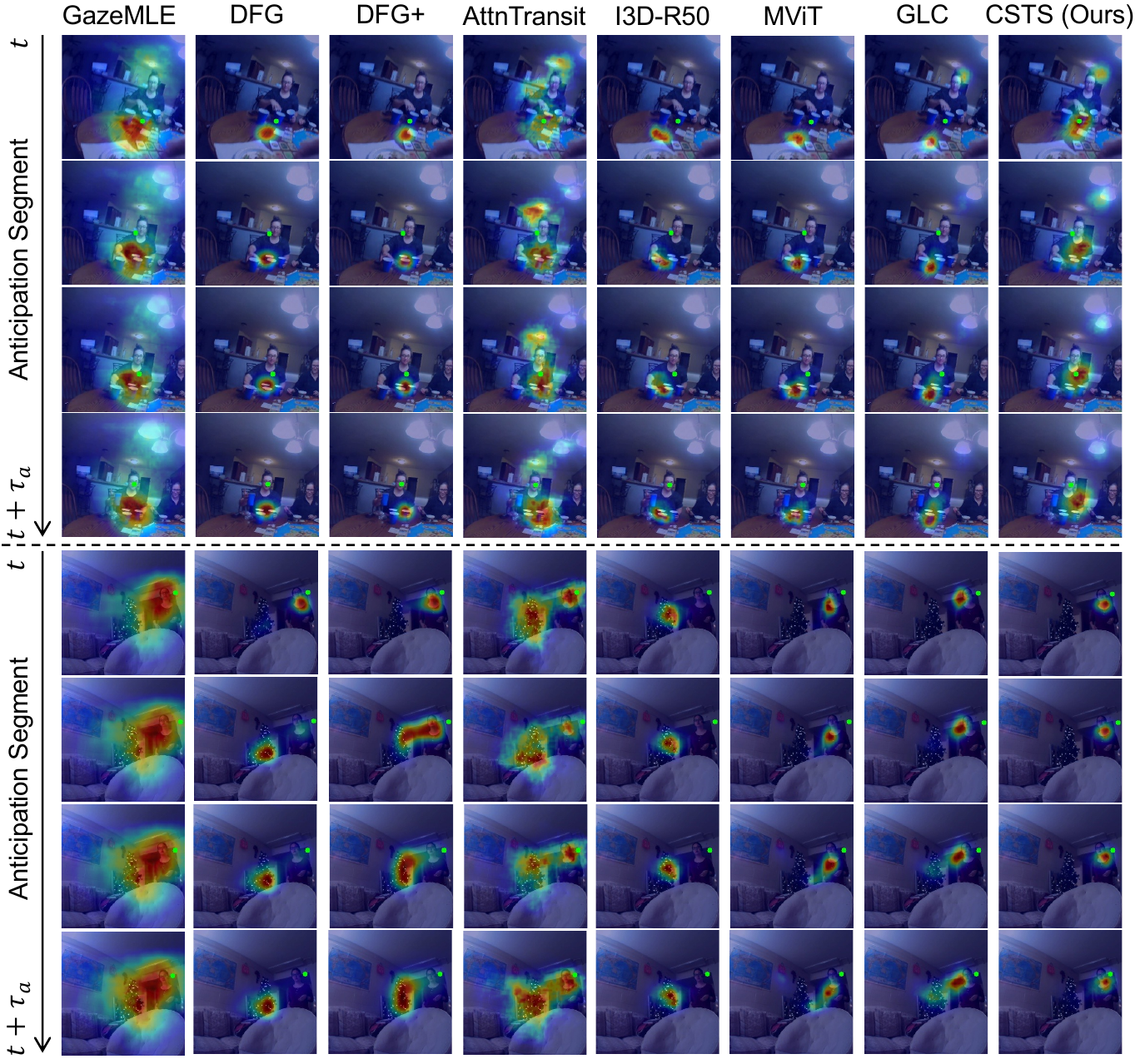}
\end{center}
\vspace{-0.3cm}
\caption{Egocentric gaze anticipation results from our model and other baselines. We show the results of four future time steps uniformly sampled from the anticipation segments. {\color{ForestGreen} Green dots} indicate the ground truth gaze location.}
\vspace{-0.4cm}
\label{fig:pred}
\end{figure}

In addition, we evaluate gaze anticipation on each anticipation time step independently and compare with previous methods in \cref{fig:perframe}. Unsurprisingly, the anticipation problem becomes more challenging as the anticipation time step increases farther into the future. Our CSTS method consistently outperforms all baselines at all future time steps. Moreover, we note that our model also produces new SOTA results on egocentric gaze \emph{estimation}, demonstrating the generalizability and robustness of our approach across gaze modeling tasks. We include these results in supplementary.

\begin{figure}[t]
\centering
\includegraphics[width=\linewidth]{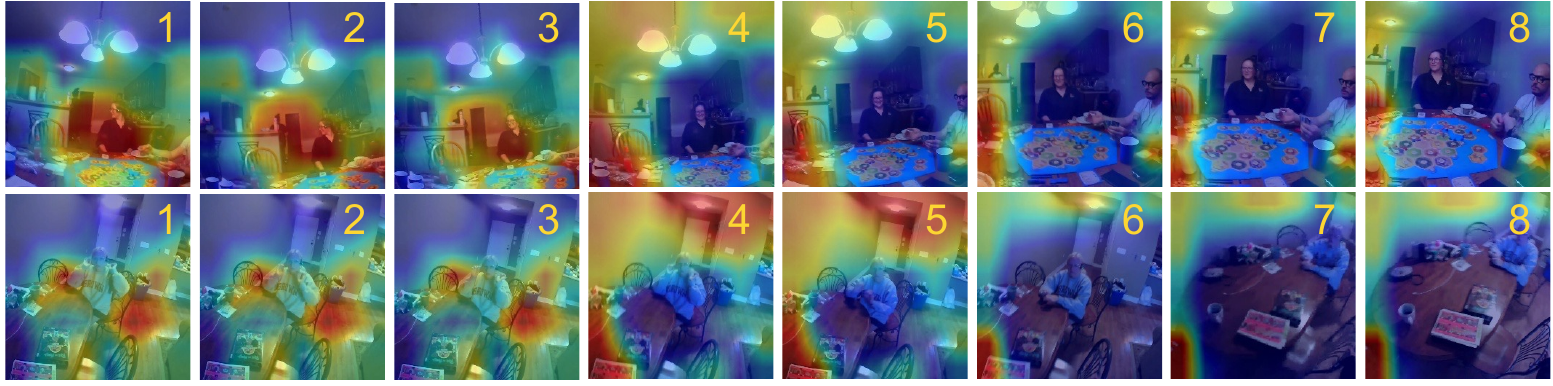}
\caption{Visualization of the spatial correlation weights. All video frames are sorted in a chronological order indexed by the numbers on the top-right corner.}
\label{fig:spatial_attn}
\vspace{-0.1cm}
\end{figure}

\subsection{Visualization of Predictions}

We visually showcase the anticipation results of CSTS and the baselines in \cref{fig:pred}. We can see that GazeMLE~\cite{li2021eye} and AttnTransit~\cite{huang2018predicting} produce more uncertainty in prediction heatmaps. Other methods fail to anticipate the true gaze target, and are likely to be misled by other salient objects. Our CSTS approach produces the best gaze anticipation results among all methods. We attribute this improvement to our novel model design that effectively addresses the unique challenges of forecasting gaze targets in egocentric view.

\subsection{Visualization of Learned Correlations}

We provide further insight on our model by visualizing the audio-visual correlations from the spatial fusion module. For each time step $t$, we calculate the correlation of each visual token with the single audio token and map it back to the input frames. The correlation heatmaps are shown in \cref{fig:spatial_attn}. In the first example, the speaker in the middle speaks, then turns her head around to talk with a social partner in the background (frame 1-3). We observe that our model captures that the audio signal has the highest correlation with spatial region of the speaker while she is speaking. Then, when she stops talking and turns her head back, the correlation is highest in the background regions, indicating the potential location of her social partner. The second example illustrates a similar phenomenon: the model captures the speaker at the beginning when she is talking, then attends to background locations when she stops. These examples suggest our model has the capability to model the audio-visual correlations in spatial dimension to learn a robust audio-visual representation.

\vspace{-0.1cm}
\section{Conclusion}
\vspace{-0.1cm}
In this paper, we propose a novel contrastive spatial-temporal separable fusion approach (CSTS) for egocentric gaze anticipation. Our key contribution is breaking down the fusion of the audio and visual modalities into a separate spatial fusion module for learning the spatial co-occurrence of visual features and audio signals, and a temporal fusion module for modeling the changing viewpoint and scene driven by audio stimuli. We further adopt a contrastive loss on the reweighted audio-visual representations from the fusion modules to facilitate multimodal representation learning. We demonstrate the benefits of our proposed model design on two egocentric video datasets: Ego4D and Aria. Our work is a key step for probing into human cognitive process with computational models, and provides important insights into multimodal representation learning, visual forecasting and egocentric video understanding.

%
%
\bibliographystyle{splncs04}
\bibliography{main}


\clearpage
\appendix
\setcounter{section}{0}

\begin{center}
    \textbf{\Large Listen to Look into the Future: Audio-Visual Egocentric Gaze Anticipation}
    \\ [0.8cm]
    {\Large Supplementary Material}
    \\ [1.2cm]
\end{center}

This is the supplementary material for the paper titled "Listen to Look into the Future: Audio-Visual Egocentric Gaze Anticipation". We organize the content as follows:

\begin{itemize}
\setlength{\itemsep}{0.22cm}
\item[$\bullet$] \textbf{\hyperref[sec:cmp_avlearning]{A} -- Comparison with Prior Audio-Visual Learning Strategies}
\item[$\bullet$] \textbf{\hyperref[sec:extra_exp]{B} -- Additional Experiment Results}
\begin{itemize}
    \setlength{\itemsep}{0.08cm}
    \item[$\diamond$] \hyperref[sec:generalization]{B.1} -- Experiments about Model Generalization Capability
    \item[$\diamond$] \hyperref[sec:estimation]{B.2} -- Experiments on Egocentric Gaze Estimation
    \item[$\diamond$] \hyperref[sec:contr]{B.3} -- Additional Experiments on Contrastive Learning
    \item[$\diamond$] \hyperref[sec:visualization]{B.4} -- Additional Visualization
\end{itemize}
\item[$\bullet$] \textbf{\hyperref[sec:implementation]{C} -- More Implementation Details}
\begin{itemize}
    \setlength{\itemsep}{0.08cm}
    \item[$\diamond$] \hyperref[sec:implementation_our_model]{C.1} -- Implementation Details of Our Model
    \item[$\diamond$] \hyperref[sec:implementation_fusion]{C.2} -- Implementation Details of Baseline Fusion Strategies
\end{itemize}
\item[$\bullet$] \textbf{\hyperref[sec:future]{D} -- Limitation and Future Work}
\item[$\bullet$] \textbf{\hyperref[sec:license]{E} -- Code and License}
\end{itemize}

\renewcommand\thesection{\Alph {section}}
\renewcommand\thesubsection{\thesection.\arabic{subsection}}

\section{Comparison with Prior Audio-Visual Learning Strategies}
\label{sec:cmp_avlearning}
We have specified the key differences between the egocentric action anticipation task and saliency prediction task in the second paragraph of \cref{sec:related_work} in the main paper. The experiment results also validate that our proposed spatial-temporal separable fusion strategy performs better in our task than other fusion strategies designed for saliency prediction and action recognition (please refer to \cref{tab:other_fusion} in the main paper). In this section, we further compare our model with typical audio-visual learning methods for saliency prediction and recognition tasks in terms of model design.

In \cref{tab:cmp_prior_methods}, all prior methods are designed for exocentric videos (\ie, third-person videos) that have a \emph{fixed} camera viewpoint through all frames. Though various fusion approaches are used in these methods, they fuse audio-visual embeddings \emph{jointly} in time and space. In contrast, the egocentric gaze anticipation task has the unique challenges of \emph{moving} viewpoint together the \emph{latency} between the audio stimuli and human reactions. To address these challenges, our model uses a novel spatial-temporal separable fusion strategy which has not been studied in prior work. The experiments in \cref{tab:other_fusion} of the main paper shows that our method achieves the best performance in egocentric gaze anticipation task compared with prior audio-visual learning strategies. In addition, using contrastive learning to boost audio-visual representations in a specific task is still an understudied area. Huang \etal~\cite{huang2024mavil} use inter- and intra- contrastive loss to learn aligned audio and visual embeddings. However, they straightforwardly apply contrastive loss on the \emph{raw} embeddings right after the encoders. In our model, we innovatively propose to adopt contrastive loss on the embeddings after fusion layers (\ie, post-fusion contrastive learning). We also validate its advantage in \cref{tab:contr} of the main paper. These key differences consolidate our contributions and clearly distinguish our model from other audio-visual learning methods.

\begin{table}[t]
\scriptsize
\centering
\setlength{\tabcolsep}{3pt}
\renewcommand\arraystretch{1.1}
\caption{Comparison with typical audio-visual learning methods for audio-visual saliency prediction and recognition. If more than one fusion strategies have been tried in one method, we only show the strategy leading to the best performance.}
\begin{tabular}{lccccc}
\toprule
Methods & View  & Fusion  & Cntr  & Architecture  & Task \\
\midrule
Tavakoli \etal~\cite{tavakoli2019dave}   & Exo  & Concatenation         & w/o  & CNN  & Saliency Prediction \\
Min \etal~\cite{min2020multimodal}       & Exo  & Correlation Analysis  & w/o  & CNN  & Saliency Prediction \\
Tsiami \etal~\cite{tsiami2020stavis}     & Exo  & Bilinear              & w/o  & CNN  & Saliency Prediction \\
Yao \etal~\cite{yao2021deep}             & Exo  & Inner Product         & w/o  & CNN  & Saliency Prediction \\
Change \etal~\cite{chang2021temporal}    & Exo  & Bilinear              & w/o  & CNN  & Saliency Prediction \\
Jain \etal~\cite{jain2021vinet}          & Exo  & Bilinear              & w/o  & CNN  & Saliency Prediction \\
Wang \etal~\cite{wang2021semantic}       & Exo  & Concatenation         & w/o  & CNN  & Saliency Prediction \\
Xiong \etal~\cite{xiong2023casp}         & Exo  & Self-Attentation  & w/o  & CNN  & Saliency Prediction \\
\midrule
Nagrani \etal~\cite{nagrani2021attention}   & Exo  & Attention Bottleneck  & w/o  & Transformer  & Video Classification \\
Huang \etal~\cite{huang2024mavil}           & Exo  & Self-Attention   & w/   & Transformer  & Video Classification \\
Gao \etal~\cite{gao2018learning}            & Exo  & Linear           & w/o  & LSTM         & Action Recognition \\
Kazakos \etal~\cite{kazakos2019epic}        & Exo  & Linear           & w/o  & CNN          & Action Recognition \\
Wang \etal~\cite{wang2020makes}             & Exo  & Weigted Sum      & w/o  & CNN          & Action Recognition \\ 
Xiao \etal~\cite{xiao2020audiovisual}       & Exo  & Self-Attentation & w/o  & CNN          & Action Recognition \\
Liu \etal~\cite{liu2023self}                & Exo  & Linear           & w/o     & CNN          & Action Recognition \\
Senocak \etal~\cite{senocak2023event}       & Exo  & Linear           & w/o     & CNN          & Action Recognition \\
Praveen \etal~\cite{praveen2022joint}       & Exo  & Self-Attention   & w/o     & CNN          & Emotion Recognition \\
Chudasama \etal~\cite{chudasama2022m2fnet}  & Exo  & Self-Attention   & w/o  & Transformer  & Emotion Recognition \\
\midrule
\multirow{2}{*}{\textbf{CSTS (Ours)}}        & \multirow{2}{*}{\textbf{Ego}}  & \textbf{Spatial-Temporal}  & \multirow{2}{*}{\textbf{w/}} & \multirow{2}{*}{\textbf{Transformer}} & \multirow{2}{*}{\textbf{Gaze Anticipation}} \\
& & \textbf{Separable}  & & & \\
\bottomrule
\end{tabular}
\label{tab:cmp_prior_methods}
\end{table}

\section{Additional Experiment Results}
\label{sec:extra_exp}

\begin{table}[t]
\centering
\caption{Zero-shot experiments on Aria dataset. All baselines and our model are trained only on Ego4D training set. We consider F1 score as the \emph{primary} metric in our experiments. The {\textcolor{ForestGreen}{green}} row refers to our model, and the best results are highlighted with \textbf{boldface}. See \cref{sec:generalization} for further discussion.}
\begin{tabular}{lccc}
\toprule
\makebox[0.2\textwidth][l]{Methods} & \makebox[0.13\textwidth][c]{F1 Score} & \makebox[0.13\textwidth][c]{Recall} & \makebox[0.13\textwidth][c]{Precision} \\
\midrule
GazeMLE~\cite{li2021eye}                  & 44.0 & 59.0 & 35.0 \\
AttnTransit~\cite{huang2018predicting}    & 43.1 & 57.5 & 34.5 \\
I3D-R50~\cite{feichtenhofer2019slowfast}  & 41.5 & 77.2 & 28.4 \\
MViT~\cite{fan2021multiscale}             & 44.1 & 59.7 & 35.0 \\
GLC~\cite{lai2022eye}                     & 46.9 & 72.8 & 34.6 \\
\hdashline
DFG~\cite{zhang2017deep}                  & 39.3 & \textbf{80.4} & 26.0 \\
DFG+~\cite{zhang2018anticipating}         & 43.1 & 76.4 & 30.0 \\
\rowcolor[HTML]{cef8d1} CSTS              & \textbf{50.8} & 62.2 & \textbf{42.9} \\
\bottomrule
\end{tabular}
\label{tab:zero_shot}
\end{table}

\subsection{Experiments about Model Generalization Capability}
\label{sec:generalization}
To validate the generalization capability of our model, we compare our model with prior state-of-the-art models in a zero-shot setting. Specifically, We train our model and all baselines with Ego4D training set and test them with Aria test set. Note that the Aria data is invisible to all models during training. The results are presented in \cref{tab:zero_shot}. Our model outperforms the best egocentric gaze anticipation model (DFG+) by +7.7\% and also exceeds the strongest baseline (GLC) by +3.9\% in F1 score (primary metric). The remarkable improvement suggests that, with our novel fusion and contrastive learning approaches, our model is able to generalize better to other unseen data, which is critical for applying it to real-world problems.

\begin{table}[t]
\centering
\caption{Comparison with prior state-of-the-art models on egocentric gaze estimation. The {\textcolor{ForestGreen}{green}} row refers to our model. The best results are highlighted with \textbf{boldface}. See \cref{sec:estimation} for further discussion.}
\begin{tabular}{lcccccc}
\toprule
\makebox[0.18\textwidth][l]{\multirow{2}{*}{Methods}}  & \multicolumn{3}{c}{Ego4D} & \multicolumn{3}{c}{Aria} \\
\cmidrule(lr){2-4} \cmidrule(lr){5-7}
& \makebox[0.11\textwidth][c]{F1 Score} & \makebox[0.11\textwidth][c]{Recall} & \makebox[0.11\textwidth][c]{Precision} & \makebox[0.11\textwidth][c]{F1 Score} & \makebox[0.11\textwidth][c]{Recall} & \makebox[0.1\textwidth][c]{Precision} \\
\midrule
Center Prior & 14.9 & 21.9 & 11.3 & 28.9 & 21.7 & 43.1 \\
GazeMLE~\cite{li2021eye} & 35.4 & 49.7 & 27.5 & 58.7 & 63.4 & 54.7 \\
AttnTransit~\cite{huang2018predicting} & 36.4 & 47.6 & 29.5 & 59.2 & 60.2 & 58.3 \\
I3D-R50~\cite{feichtenhofer2019slowfast} & 37.5 & 52.5 & 29.2 & 60.9 & 69.5 & 54.2 \\
MViT~\cite{fan2021multiscale} & 40.9 & 57.4 & 31.7 & 61.7 & \textbf{71.2} & 54.5 \\
GLC~\cite{lai2022eye} & 43.1 & 57.0 & 34.7 & 63.2 & 67.4 & 59.5 \\
\rowcolor[HTML]{cef8d1} CSTS & \textbf{43.7} & \textbf{58.0} & \textbf{35.1} & \textbf{64.5} & 69.6 & \textbf{60.1} \\
\bottomrule
\end{tabular}
\label{tab:sota_estimation}
\end{table}

\subsection{Experiments on Egocentric Gaze Estimation}
\label{sec:estimation}
In addition to egocentric gaze anticipation, we also evaluate the advantage of our model in another gaze modeling problem -- egocentric gaze estimation. Instead of forecasting \emph{future} gaze, egocentric gaze estimation requires gaze target prediction in the \emph{current} video frames. We use the same experiment setup from the recent state-of-the-art method~\cite{lai2022eye}.

As demonstrated in \cref{tab:sota_estimation}, the prior work~\cite{lai2022eye} has shown the superiority of using a transformer-based architecture for egocentric gaze estimation. By incorporating the audio modality, CSTS surpasses the backbone MViT~\cite{fan2021multiscale} (vision-only counterpart) by +2.8\% on both Ego4D on Aria in terms of F1 score. These results indicate the audio modality also makes important contributions to the performance on egocentric gaze estimation. Furthermore, our model outperforms GLC~\cite{lai2022eye} by +0.6\% and +1.3\% on Ego4D and Aria respectively, achieving a new state-of-the-art performance for this problem. However, our method has a smaller performance improvement on the gaze estimation task compared to gaze anticipation. The possible reason is that the audio stream has a stronger connection with future gaze targets than current gaze behaviors because of the natural latency between the audio stimuli and human reactions.

\subsection{Additional Experiments on Contrastive Learning}
\label{sec:contr}

\begin{table}[t]
\centering
\caption{Study of different strategies for contrastive loss implementation. Post Cntr refers to our proposed post-fusion contrastive learning strategy, and the {\textcolor{ForestGreen}{green}} row refers to the complete CSTS model. The best results are highlighted with \textbf{boldface}. See \cref{sec:contr} for further discussion.}
\begin{tabular}{lcccccc}
\toprule
\makebox[0.26\textwidth][l]{\multirow{2}{*}{Methods}}  & \multicolumn{3}{c}{Ego4D} & \multicolumn{3}{c}{Aria} \\
\cmidrule(lr){2-4} \cmidrule(lr){5-7}
& \makebox[0.11\textwidth][c]{F1 Score} & \makebox[0.11\textwidth][c]{Recall} & \makebox[0.11\textwidth][c]{Precision} & \makebox[0.11\textwidth][c]{F1 Score} & \makebox[0.11\textwidth][c]{Recall} & \makebox[0.11\textwidth][c]{Precision} \\
\midrule
STS + Vanilla Contr & 39.0 & 53.7 & 30.6 & 59.1 & 66.5 & 53.1 \\
STS + S-Contr & 38.5 & 53.5 & 30.0 & 59.0 & 66.3 & 53.1 \\
STS + T-Contr & 38.9 & 54.0 & 30.5 & 59.0 & 66.7 & 53.0 \\
STS + Cross Contr & 38.9 & \textbf{54.4} & 30.2 & 59.3 & 66.8 & 53.3  \\
\rowcolor[HTML]{cef8d1} STS + Post Contr & \textbf{39.7} & 53.3 & \textbf{31.6} & \textbf{59.9} & \textbf{66.8} & \textbf{54.3} \\
\bottomrule
\end{tabular}
\label{tab:supp_contr}
\end{table}

In our model, we propose to use the audio-visual representations obtained after fusion (\ie $u_v$ and $u_a$) to calculate contrastive loss (\ie, post-fusion contrastive learning). As a comparison, we also implement a baseline by feeding the raw embeddings from the encoders (\ie $\phi(x)$ and $\psi(a)$) to the contrastive loss which is denoted as \textbf{Vanilla Contr}. To further investigate the contribution of contrastive learning, we also conduct experiments with three additional strategies:

\noindent\textbf{Cross Contr}.\ In our final model (CSTS), we use the new visual representation $u_v=u_{v,s}\otimes u_{v,t}$ and the new audio representation $u_a=\psi(a)\otimes u_{a,t}$ as input to the contrastive loss. In Cross Contr, we still use $u_v$ yet replace $u_a$ by reweighting the audio representation $u_{a,s}$ after the spatial fusion with weight $u_{a,t}$ from the temporal fusion, \ie $u_a^*=u_{a,s}\otimes u_{a,t}$, as input to the contrastive loss. Please refer to \cref{fig:overview} in the main paper for the meaning of each notation.

\noindent\textbf{S-Contr}.\ We use the output from the spatial fusion module ($u_{v,s}$,$u_{a,s}$) to calculate the contrastive loss. 

\noindent\textbf{T-Contr}.\ We use the output from the temporal fusion module ($u_{v,t}$,$u_{a,t}$) to calculate the contrastive loss. 

We implement all contrastive learning baselines above on our proposed model architecture and fusion strategy (\ie, STS). The results are summarized in \cref{tab:supp_contr}. Both S-Contr and T-Contr lag behind or perform on par with Vanilla Contr. One possible reason is that conducting contrastive learning using features obtained from only one fusion branch may compromise the representation learning of the other branch. Additionally, Cross Contr works on-par with Vanilla Contr on Ego4D but performs better on Aria. It also consistently outperforms S-Contr and T-Contr. This result validates our claim that implementing contrastive loss with reweighted representations from both spatial and temporal fusion leads to more gains for egocentric gaze anticipation. Moreover, our proposed strategy (reweighting the raw audio embedding $\psi(a)$ rather than the fused embedding after spatial fusion) outperforms Cross Contr. This is because in Cross Contr $u_{a,s}$ is derived from spatial fusion, where each audio token is fused with 64 visual tokens in the spatial fusion branch resulting in the dilution of audio features. All results further demonstrate the benefits of our proposed contrastive learning strategy.

\begin{figure}[t]
\begin{center}
\includegraphics[width=0.86\linewidth]{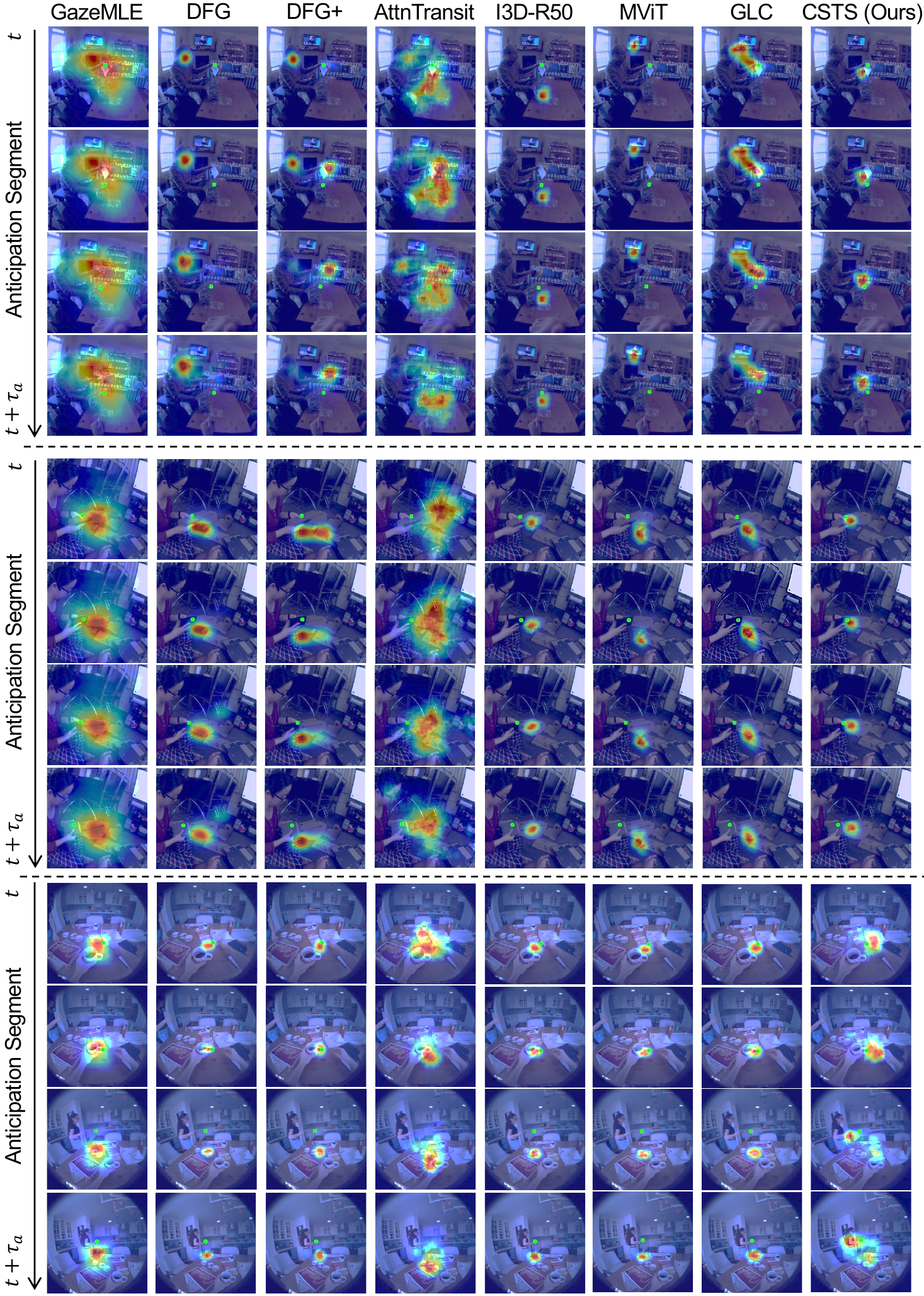}
\end{center}
\vspace{-1.2em}
\caption{Additional egocentric gaze anticipation results from our model and other baselines. {\color{ForestGreen} Green dots} indicate the ground truth gaze location. The first two examples are from the Ego4D dataset, and the last example is from the Aria dataset.}
\label{fig:pred_supp}
\end{figure}

\begin{figure}[t]
\begin{center}
\includegraphics[width=0.86\linewidth]{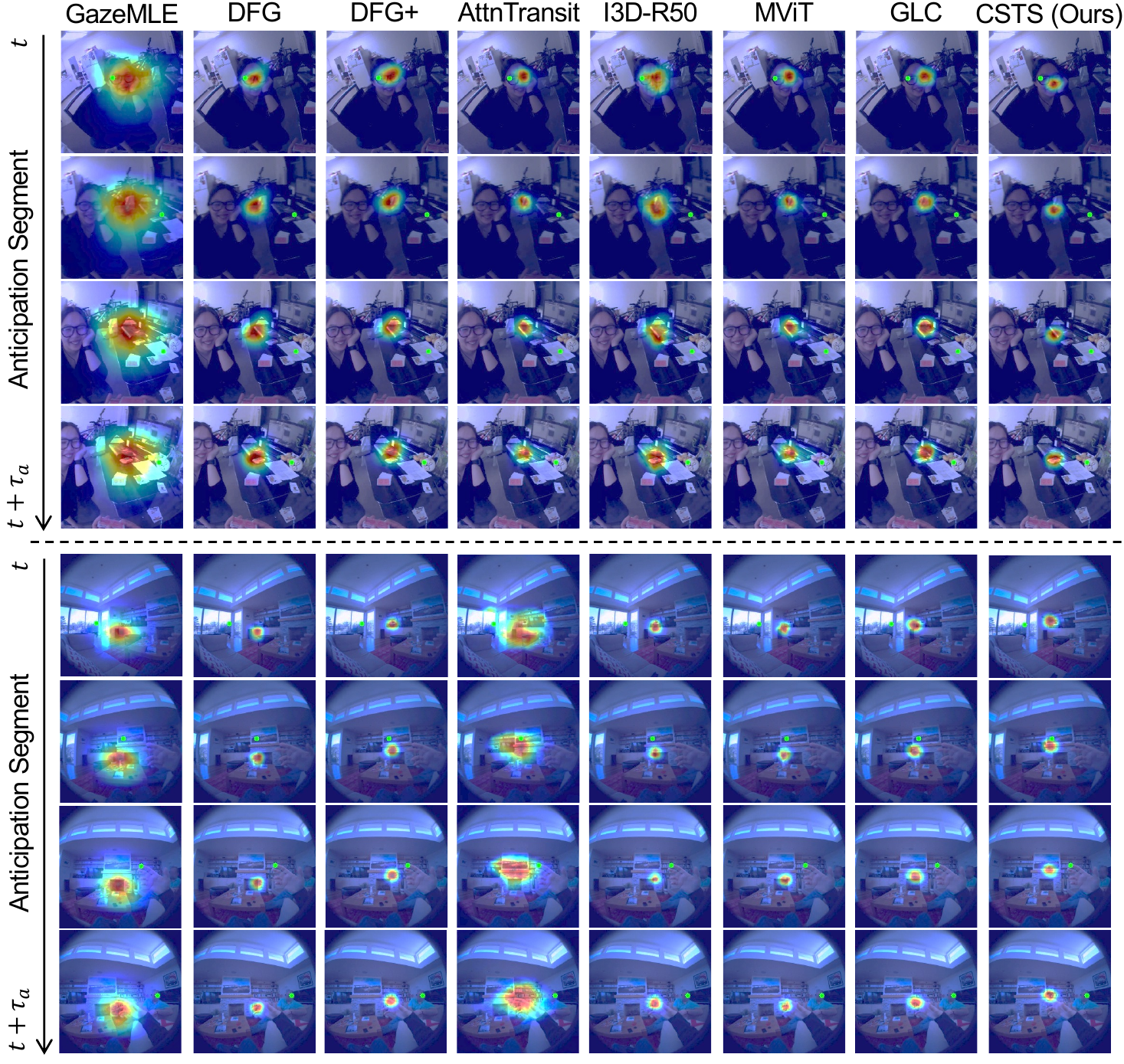}
\end{center}
\caption{Failure cases of our model and baselines. {\color{ForestGreen} Green dots} indicate the ground truth gaze location. The first example is from the Ego4D dataset, and the second example is from the Aria dataset.}
\label{fig:failure}
\end{figure}

\subsection{Additional Visualization}
\label{sec:visualization}
We showcase more qualitative comparisons with all the baselines for egocentric gaze anticipation in \cref{fig:pred_supp}. We observe CSTS makes the most accurate predictions. We also illustrate some typical failure cases in \cref{fig:failure}. In the first example, our model makes an accurate prediction in the first frame but fails at the following time steps due to the gaze movement. In the second example, the the camera view and gaze target move from the left to the right. This drastic change causes the mistake in our model's predictions. Similar failures also happen in the predictions of all baselines. Notably, existing deep models tend to only successfully anticipate steady gaze fixations or small gaze movements in the near future, and can not effectively capture large gaze shifts. This is the a common limitation shared by many existing works of future anticipation~\cite{jia2022generative} in egocentric videos.

\section{More Implementation Details}
\label{sec:implementation}

\subsection{Implementation Details of Our Model}
\label{sec:implementation_our_model}
\noindent\textbf{Architecture}.\ Inspired by~\cite{gao2020listen}, we use a light-weight audio encoder composed of four self-attention blocks from MViT~\cite{fan2021multiscale}. The model architecture is further detailed in \cref{tab:architecture}. We initialize the video encoder with Kinetics-400 pretraining~\cite{kay2017kinetics} and initialize the audio encoder using Xavier initialization~\cite{glorot2010understanding}. The resulting video embeddings $\phi(x)$ have a dimension of $T=4, H=8, W=8, D=768$, and the resulting audio embeddings $\psi(a)$ have a dimension of $T=4, M=64, D=768$. We follow~\cite{lin2022egocentric} to map audio-visual representation vectors to dimension $D'=256$ for the contrastive loss. The output from the decoder is a downsampled heatmap which is upsampled to match the input size using trilinear interpolation. Following~\cite{lai2022eye}, we add intermediate features from each video encoder block to the corresponding decoder block output via skip connections to compensate for the loss of low-level textures.

\noindent\textbf{Training}.\ We set both temperature factor $\mathcal{T}$ of contrastive loss and re-weight parameter $\alpha$ as 0.05. Follow~\cite{li2021eye,lai2022eye}, we use a Gaussian distribution with kernel size of 19 centered on the gaze location in each frame as the ground truth gaze heatmap during training. The model is trained with AdamW~\cite{loshchilov2017decoupled} optimization for 15 epochs. The momentum and weight decay are set as 0.9 and 0.05. The initial learning rate is $10^{-4}$ which decreases with the cosine learning rate decay strategy. The model is trained with a batch size of 8 across 4 GPUs.

\subsection{Implementation Details of Baseline Fusion Strategies}
\label{sec:implementation_fusion}
We compare with multiple different audio-visual fusion strategies in main paper \cref{tab:other_fusion}. The details of each baseline are listed as follows:

\noindent\textbf{Linear.} We reshape the video embedding and audio embedding to the shape $\hat{N}\times D$. We concatenate the two reshaped embeddings (resulting in the dimension of $\hat{N}\times2D$) and input it to two linear layers. The dimension of the output is $\hat{N}\times D$ and we reshape it back to $T\times H\times W\times D$ which is fed into the decoder.

\noindent\textbf{Bilinear.} We reduce the length of video tokens and audio tokens to 256 using a linear layer respectively. Then we input the resulting video and audio tokens into a bilinear layer. The output is fed into the decoder for gaze forecasting.

\noindent\textbf{Concat.} We reshape the audio embedding $\psi(a)\in\mathbb{R}^{T\times H\times W\times D}$ to the same dimension as the video embedding $\phi(x)\in\mathbb{R}^{T\times H\times W\times D}$ and concatenate them along the channel to obtain an audio-visual representation with dimension of $T\times H\times W\times 2D$. This representation is fed into the decoder for gaze forecasting.

\noindent\textbf{Vanilla SA.} In this baseline, we flatten the video embedding and audio embedding into a list of tokens and thereby obtain $T \times (N+M)$ tokens in total, where $N=H\times W$. Then we input all tokens to a standard self-attention layer followed by multiple linear layers to perform fusion in the spatial and temporal dimensions simultaneously. We split the output into a new visual embedding incorporating audio information with dimension of $T\times N\times D$ and a new audio embedding incorporating visual information with dimension of $T\times M\times D$. The new visual embedding is input into the decoder.

\noindent\textbf{STS}.\ This is a baseline using the same fusion strategy as our method but without using the contrastive loss for training.

\section{Limitation and Future Work}
\label{sec:future}
In this paper, we propose a novel contrastive spatial-temporal separable fusion model for audio-visual egocentric gaze anticipation. Our method is validated on the Ego4D~\cite{grauman2022ego4d} and Aria~\cite{aria_pilot_dataset} datasets. Our method has larger performance improvement on the Aria dataset comparing with Ego4D dataset. We believe this is because the multi-person social interaction setting from Ego4D dataset incurs additional challenges for audio representation learning, like multiple people and speakers present. Our current model design did not explicitly address this challenging nature of multi-speaker social interactions. Another limitation is that our model fails to anticipate the drastic gaze movements (see the failure cases in \cref{fig:failure}). In addition, in this work we do not explore the spatial geometry context provided by multi-channel audio signals. Our approach and experiments sugggest several important future research directions:
\begin{itemize}
    \item[$\bullet$] The proposed CSTS model can be applied to other video understanding tasks related to the audio modality, such as action recognition, action localization, and video question answering. We hope to further investigate our proposed approach on these problem settings.

    \item[$\bullet$] A model explicitly designed for audio-visual representation learning in multi-person, multi-speaker environments merits further investigation.

    \item[$\bullet$] A model that learns better temporal representations for anticipating large gaze shifts remains to be explored.

    \item[$\bullet$] The visualization of correlation weights in the spatial fusion module indicates the potential of our model for weakly-supervised/self-supervised sound localization and active speaker detection, which can be investigated in further work.
\end{itemize}

\section{Code and License}
\label{sec:license}
The usage of the Aria dataset is under the Apache 2.0 License$^1$\footnote{1\quad\url{https://github.com/facebookresearch/vrs/blob/main/LICENSE}}, and the usage of the Ego4D dataset is under the license agreement$^2$\footnote{2\quad\url{https://ego4d-data.org/pdfs/Ego4D-Licenses-Draft.pdf}}. Our implementation is built on top of~\cite{fan2020pyslowfast}, which is under the Apache License$^3$\footnote{3\quad\url{https://github.com/facebookresearch/SlowFast/blob/main/LICENSE}}. Our code and the train/test split on Aria dataset will be available at: \url{https://bolinlai.github.io/CSTS-EgoGazeAnticipation/}.

\begin{table}[t]
\scriptsize
\begin{center}
\renewcommand\arraystretch{1.2}
\begin{tabular}{c|c|c|c}
\hline
 & Stages & Operators & Output Size \\
\hline
\multirow{12}{*}{\rotatebox[origin=c]{90}{Video Encoder $\phi(x)$}}  
& video frames  & -  & $8\times256\times256\times3$ \\
\cline{2-4}
& video token embedding & $\begin{array}{c} Conv(3\times7\times7,\ 96) \\ stride\ 2\times4\times4 \end{array}$ & $4\times64\times64\times96$ \\
\cline{2-4}
& tokenization & flattening & $(4\times64\times64)\times96$ \\
\cline{2-4}
& video encoder block1          & $\left[\begin{array}{c} MSA(96) \\ MLP(384)\end{array}\right]\times1$ & $(4\times64\times64)\times192$ \\
\cline{2-4}
& video encoder block2          & $\left[\begin{array}{c} MSA(192) \\ MLP(768)\end{array}\right]\times2$ & $(4\times32\times32)\times384$ \\
\cline{2-4}
& video encoder block3          & $\left[\begin{array}{c} MSA(384) \\ MLP(1536)\end{array}\right]\times11$ & $(4\times16\times16)\times768$ \\
\cline{2-4}
& video encoder block4          & $\left[\begin{array}{c} MSA(768) \\ MLP(3072)\end{array}\right]\times2$ & $(4\times8\times8)\times768$ \\
\hline
\multirow{12}{*}{\rotatebox[origin=c]{90}{Audio Encoder $\psi(a)$}}  
& audio spectrograms  & -  & $8\times256\times256\times1$ \\
\cline{2-4}
& audio token embedding & $\begin{array}{c} Conv(3\times7\times7,\ 96) \\ stride\ 2\times4\times4 \end{array}$ & $4\times64\times64\times96$ \\
\cline{2-4}
& tokenization & flattening & $(4\times64\times64)\times96$ \\
\cline{2-4}
& audio encoder block1          & $\left[\begin{array}{c} MSA(96) \\ MLP(384)\end{array}\right]\times1$ & $(4\times4096)\times192$ \\
\cline{2-4}
& audio encoder block2          & $\left[\begin{array}{c} MSA(192) \\ MLP(768)\end{array}\right]\times1$ & $(4\times1024)\times384$ \\
\cline{2-4}
& audio encoder block3          & $\left[\begin{array}{c} MSA(384) \\ MLP(1536)\end{array}\right]\times1$ & $(4\times256)\times768$ \\
\cline{2-4}
& audio encoder block4          & $\left[\begin{array}{c} MSA(768) \\ MLP(3072)\end{array}\right]\times1$ & $(4\times64)\times768$ \\
\hline
\multirow{12}{*}{\rotatebox[origin=c]{90}{Fusion Modules}}
& conv1  & $\begin{array}{c} Conv(768\times1\times8\times8,\ 768) \\ stride\ 1\times1\times1 \end{array}$ & $4\times1\times768$ \\
\cline{2-4}
& in-frame self-attention $\sigma(\cdot)$          & $\left[\begin{array}{c} MSA(768) \\ MLP(3072)\end{array}\right]\times1$ & $4\times(64+1)\times768$ \\
\cline{2-4}
& conv2  & $\begin{array}{c} Conv(768\times1\times8\times8,\ 768) \\ stride\ 1\times1\times1 \end{array}$ & $4\times1\times768$ \\
\cline{2-4}
& conv3  & $\begin{array}{c} Conv(768\times1\times8\times8,\ 768) \\ stride\ 1\times1\times1 \end{array}$ & $4\times1\times768$ \\
\cline{2-4}
& cross-frame self-attention $\pi(\cdot)$         & $\left[\begin{array}{c} MSA(768) \\ MLP(3072)\end{array}\right]\times1$ & $8\times1\times768$ \\
\cline{2-4}
& reweighting & $ u_{v,s}\otimes u_{v,t}$ & $8\times64\times768$ \\
\cline{2-4}
& reweighting & $\psi(a)\otimes u_{a,t}$ & $8\times64\times768$ \\
\hline
\multirow{10}{*}{\rotatebox[origin=c]{90}{Decoder}}   & decoder block1          & $\left[\begin{array}{c} MSA(1536) \\ MLP(3072)\end{array}\right]\times1$ & $(4\times16\times16)\times768$ \\
\cline{2-4}
& decoder block2          & $\left[\begin{array}{c} MSA(768) \\ MLP(1536)\end{array}\right]\times1$ & $(4\times32\times32)\times384$ \\
\cline{2-4}
& decoder block3          & $\left[\begin{array}{c} MSA(384) \\ MLP(768)\end{array}\right]\times1$ & $(4\times64\times64)\times192$ \\
\cline{2-4}
& decoder block4          & $\left[\begin{array}{c} MSA(192) \\ MLP(384)\end{array}\right]\times1$ & $(8\times64\times64)\times96$ \\
\cline{2-4}
& head                    & $\begin{array}{c} Conv(1\times1\times1,\ 1) \\ stride\ 1\times1\times1 \end{array}$ & $8\times64\times64\times1$ \\
\hline
\end{tabular}
\end{center}
\caption{Architecture of the proposed model. Convolutional layers are denoted as $Conv(kernel\ size,\ output\ channels)$. The number of input channels in multi-head self-attention is shown in the parenthesis of $MSA$. The dimension of the hidden layer in multi-layer perceptron is listed in parenthesis of $MLP$. conv1 is the convolutional layer in the spatial fusion module. conv2 and conv3 are convolutional layers in the temporal fusion module.}
\label{tab:architecture}
\end{table}

\end{document}